\documentclass[pdflatex,sn-mathphys-num]{sn-jnl}% Math and Physical Sciences Numbered Reference Style
\usepackage{graphicx}%
\usepackage{multirow}%
\usepackage{amsmath,amssymb,amsfonts}%
\usepackage{amsthm}%
\usepackage{mathrsfs}%
\usepackage[title]{appendix}%
\usepackage{xcolor}%
\usepackage{textcomp}%
\usepackage{manyfoot}%
\usepackage{booktabs}%
\usepackage{algorithm}%
\usepackage{algorithmicx}%
\usepackage{algpseudocode}%
\usepackage{listings}%
\usepackage{subcaption}%
\usepackage{tikz}%

\usepackage[T1]{fontenc}
\usepackage[utf8]{inputenc}

\definecolor{viola}{HTML}{4D2FB2}

\theoremstyle{thmstyleone}%
\theoremstyle{thmstyletwo}%

\theoremstyle{thmstylethree}%

\begin{document}

\title[The Statistical Signature of LLMs]{The Statistical Signature of LLMs}


\author[1]{\fnm{Ortal} \sur{Hadad}}

\author[2]{\fnm{Edoardo} \sur{Loru}}

\author[1]{\fnm{Jacopo} \sur{Nudo}}

\author[3]{\fnm{Niccolò} \sur{Di Marco}}

\author[1]{\fnm{Matteo} \sur{Cinelli}}

\author*[1]{\fnm{Walter} \sur{Quattrociocchi}}\email{walter.quattrociocchi@uniroma1.com}

\affil[1]{Department of Computer Science, Sapienza University of Rome}

\affil[2]{Department of Computer, Control and Management Engineering, Sapienza University of Rome}

\affil[3]{Department of Legal, Social, and Educational Sciences, Tuscia University}

\abstract{Large language models generate text through probabilistic sampling from high-dimensional distributions, yet how this process reshapes the structural statistical organization of language remains incompletely characterized. Here we show that lossless compression provides a simple, model-agnostic measure of statistical regularity that differentiates generative regimes directly from surface text.

We analyze compression behavior across three progressively more complex information ecosystems: controlled human–LLM continuations, generative mediation of a knowledge infrastructure (Wikipedia vs. Grokipedia), and fully synthetic social interaction environments (Moltbook vs. Reddit). Across settings, compression reveals a persistent structural signature of probabilistic generation. In controlled and mediated contexts, LLM-produced language exhibits higher structural regularity and compressibility than human-written text, consistent with a concentration of output within highly recurrent statistical patterns. However, this signature shows scale dependence: in fragmented interaction environments the separation attenuates, suggesting a fundamental limit to surface-level distinguishability at small scales.

This compressibility-based separation emerges consistently across models, tasks, and domains and can be observed directly from surface text without relying on model internals or semantic evaluation. Overall, our findings introduce a simple and robust framework for quantifying how generative systems reshape textual production, offering a structural perspective on the evolving complexity of communication.
}

\keywords{large language models, stochastic generation, lossless compression, synthetic text, information dynamics}



\maketitle

\section{Introduction}\label{sec:intro}

The rapid diffusion of large language models (LLMs) is reshaping how language is produced, accessed, and circulated \cite{burton2024large}. Generative systems are now embedded in search engines, productivity tools, knowledge platforms, educational settings, and social media, increasingly mediating everyday interactions with information \cite{wessel2025generative}. As a result, large volumes of online text are partially or entirely generated by probabilistic models, often without clear signaling or stable boundaries between human and synthetic content \cite{dugan2023real,wu2025survey}. This shift constitutes not only a technical challenge, but a structural transformation of information ecosystems, with implications for accountability, governance, and epistemic literacy \citep{bender2021stochastic,quattrociocchi2025faultlines}.

At the surface level, LLM outputs often appear coherent and stylistically similar to human writing \cite{crothers2023machine}. This resemblance has fueled, on the one hand, inflated claims about near-human linguistic competence \cite{duncan2024does}, and on the other, a large literature on detection, labeling, provenance, and watermarking of synthetic text \citep{weberwulff2023testing,perkins2024bypass,mitchell2023detectgpt,kirchenbauer2023watermark}. Empirical evidence, however, shows that detection remains fragile in open-world settings, with performance degrading under domain shift, paraphrasing, translation, post-editing, and changes in model families \citep{weberwulff2023testing,perkins2024bypass}. This brittleness suggests that the central issue is not classifier accuracy per se, but limited understanding of the structural statistical properties that differentiate language produced under distinct generative regimes.

Human language arises from complex cognitive and social processes, shaped by intention, communicative goals, pragmatic constraints, contextual grounding, and uneven information exposure across individuals and communities \cite{pinker2013language,ellis2019essentials,DiMarco2024}. LLM-generated text, by contrast, is produced through iterative sampling from conditional probability distributions learned from large corpora of prior language use \citep{vaswani2017attention,holtzman2019curious}. Even when outputs appear similar, the underlying generative mechanisms differ fundamentally. Recent work has highlighted this divergence from multiple angles. \citet{loru2025simulation} show that LLMs can replicate human-like judgment outputs while relying on qualitatively different evaluative strategies, leading to surface alignment that masks deeper epistemic mismatches. \citet{quattrociocchi2025faultlines} formalize these mismatches as epistemological fault lines and introduce the notion of \emph{Epistemia}, in which linguistic plausibility substitutes for epistemic evaluation. Complementarily, \citet{nudo2025generative} document \emph{generative exaggeration}, whereby probabilistic generation amplifies statistical regularities in simulated discourse, producing reduced variability relative to human language.

Information-theoretic concepts provide a useful theoretical reference for comparing generative regimes in a model-agnostic setting. Here, in particular, we adopt an approach based on lossless compression. Compression exploits redundancies in symbol sequences, yielding shorter encodings for text that exhibits stronger structural regularity. This connection is classical in information theory \citep{cover2006elements} and underlies universal compression schemes such as Lempel--Ziv \citep{ziv1977universal}. We use \texttt{gzip}, standardized in RFC~1952 \citep{deutsch1996gzip}, which operates directly on surface strings without access to model internals, semantic representations, or task-specific features, making it suitable for cross-domain comparisons in partially observable settings.

We apply this compression-based framework across three contexts of increasing realism: (i) controlled corpora where humans and LLMs produce text under comparable conditions; (ii) generative mediation of a large-scale knowledge infrastructure (Wikipedia vs.\ Grokipedia); and (iii) fully synthetic social interaction environments (Moltbook vs.\ Reddit). Compression reveals systematic differences in structural regularity between human and probabilistically generated language, as well as highlighting scale-dependent effects. Rather than proposing a detector of synthetic text or assessing semantic quality, we identify a structural signature of probabilistic language generation with implications across diverse settings.

\section{State of the Art}

The rapid diffusion of large language models (LLMs) has generated a large and growing literature on distinguishing human-authored text from machine-generated content, largely framed as a detection problem motivated by concerns over misinformation, academic integrity, and content moderation \cite{wu2023survey}. Approaches span supervised classifiers, zero-shot statistical methods, watermarking schemes, and analyses of robustness and evasion.
Supervised detection methods typically rely on discriminative models trained on labeled corpora to separate human and machine-generated text. While effective in controlled, in-domain settings, their performance often degrades under domain shift, changes in source models, and simple transformations such as paraphrasing or post-editing. Benchmark datasets such as HC3 \cite{guo2023hc3} and its extensions \cite{su2023hc3plus} have been introduced to support evaluation across multiple domains and tasks, including settings where semantic invariance makes detection substantially harder.
Parallel work has explored zero-shot signals derived from language-model likelihoods, token ranks, and related statistics. These methods exploit the tendency of common decoding strategies to concentrate probability mass on high-likelihood continuations, leaving detectable statistical traces. GLTR visualizes token-rank distributions and aggregates simple statistics across sampling regimes \cite{gehrmann2019gltr}, while related approaches use perplexity, entropy-adjacent measures, and variability indicators such as burstiness, trading simplicity and model-agnosticism for sensitivity to decoding parameters and post-processing.
More recent methods introduce theoretically grounded zero-shot criteria based on properties of the probability landscape. DetectGPT estimates local curvature of the log-probability function via stochastic perturbations, showing that machine-generated text tends to occupy regions of negative curvature \cite{mitchell2023detectgpt}. Fast-DetectGPT improves efficiency by replacing explicit perturbations with curvature proxies derived from conditional probabilities \cite{bao2024fastdetectgpt}, with further work examining robustness to perturbation strategies and token-level choices \cite{liu2024pecola}.
An alternative line of work frames detection as a comparison between generative distributions rather than as a classification task. Binoculars contrasts likelihoods assigned by closely related language models, demonstrating accurate zero-shot detection without task-specific training data \cite{hans2024binoculars}. This perspective emphasizes structural differences between generative processes rather than purely discriminative cues.
Beyond post-hoc detection, watermarking and provenance methods aim to embed detectable signals at generation time. Kirchenbauer et al.\ propose a watermarking scheme based on biased token sampling over a secret vocabulary subset \cite{kirchenbauer2023watermark}. While effective under cooperative settings, such approaches raise concerns regarding robustness, adversarial removal, and applicability to pre-existing content.
Across empirical evaluations, a recurring finding is the limited robustness of both supervised and zero-shot detectors in real-world conditions, with substantial false positives and false negatives and vulnerability to simple obfuscation strategies \cite{weberwulff2023testing,perkins2024bypass}. Consistent with these limitations, OpenAI discontinued its own AI text classifier due to insufficient accuracy \cite{openai2023classifier}.
Alongside detection-focused approaches, a smaller literature explores lightweight methods based on structural statistical features that avoid large models at inference time. Compression-based classification techniques exploit the link between statistical regularity and compressibility \cite{Jiang2023, Loru2026}. Berchtold et al.\ introduce a GZIP-KNN method combining lossless compression with nearest-neighbor classification \cite{berchtold-etal-2024-detecting}, connecting to earlier work linking compression and linguistic structure \cite{mahoney1999compression, Jiang2023}. Our work builds on this perspective by using compression not as a detector, but as a signal of statistical regularity, by examining its behavior across controlled corpora, generative mediation of knowledge infrastructures, and fully synthetic social interaction environments.

\section{Methods}

\subsection{Token distributions and controlled statistical regimes}\label{subsec:token_distr}
To examine how compression responds to statistical token regularity, we generate token sequences with varying levels of Shannon entropy.
Let $W = \{w_1, \ldots, w_n\}$ be a vocabulary and

$$
T = (t_1, \ldots, t_N), \qquad t_i \in W,
$$
a text of length $N$ composed of tokens drawn from $W$.  
The empirical token distribution of $T$ is defined as

\begin{equation}
p_T(w) = \frac{1}{N} \sum_{i=1}^{N} \mathbf{1}_{\{t_i = w\}}.
\end{equation}
where $\mathbf{1}_{\{t_i = w\}}$ is the indicator function.
Given a probability mass function $p : W \to [0,1]$, we generate synthetic sequence of tokens by independent sampling

$$
t_1, \ldots, t_N \sim p,
$$
after which, tokens are concatenated to form a text $T$.

To control the statistical concentration in an empirical token distribution, we consider a parametric family of distributions with fixed entropy indexed by a parameter $h \in \left[\frac{1}{n}, 1\right]$:

\begin{equation}
\label{eq:controlled_entropy_distribution}
p^{(h)}(w_1) = h, \qquad
p^{(h)}(w_i) = \frac{1-h}{n-1}, \quad i = 2, \ldots, n.
\end{equation}
The entropy of this distribution is:

\begin{equation}
H\big(p^{(h)}\big)=-
\left[h\log h+(1-h)\log\left(\frac{1-h}{n-1}\right)\right],
\end{equation}
and it is bounded at the extremes:

\begin{itemize}
    \item If $h = 1$, then $p^{(1)}$ is a Dirac distribution and $H\big(p^{(1)}\big) = 0$.
    \item If $h = \frac{1}{n}$, then $p^{(1/n)}$ is uniform over $W$ and $H\big(p^{(1/n)}\big) = \log n$.
\end{itemize}
Varying $h$ therefore defines a continuum of reference regimes ranging from highly uniform to strongly concentrated token distributions. These regimes provide an interpretable baseline for the empirical analyses presented in Figure~\ref{fig:entropy_vs_compression}, where compression behavior is evaluated across different levels of distributional concentration.

% \begin{figure}[!ht]
%     \centering
%     \includegraphics[width=0.5\linewidth]{figures/relation_h_vs_entropy.pdf}
%     \caption{Entropy of $p^{(h)}$ based on parameter $h$. As $h$ shifts from $\frac{1}{n}$ to $1$, the distribution changes from uniform to degenerate.}
%     \label{fig:h_vs_entropy}
% \end{figure}

\subsection{Compression-based structural regularity}

From an information-theoretic perspective, lossless compression can be interpreted as an operational probe of effective statistical regularity. Universal coding schemes such as Lempel--Ziv adapt to recurring patterns in symbol sequences without requiring access to the underlying generative process. Therefore, we use it as a model-agnostic measure of structural organization that can be applied consistently across heterogeneous text sources.

For each document $x$, we compute a compression-based measure using lossless compression. All texts are encoded in UTF-8 and processed as raw byte sequences without tokenization or normalization, ensuring that the analysis depends only on observable surface structure.

We use the \texttt{gzip} algorithm with default settings. Given a document $x$, its compressed size $C(x)$ (in bytes) defines the compression ratio

\begin{equation}
R(x) = \frac{C(x)}{|x|},
\label{eq:comp_ratio}
\end{equation}

where $|x|$ denotes the uncompressed size in bytes. Lower values of $R(x)$ indicate higher compressibility and stronger sequential regularity.

To examine how structural organization accumulates with length, we compute prefix-based compression curves. For a byte sequence $x = (b_1, \dots, b_n)$, prefixes $x_{1:k} = (b_1, \dots, b_k)$ are constructed for increasing $k$, and $R(x_{1:k})$ is evaluated independently. This procedure captures how regularity emerges as longer-range dependencies become available to the compressor.

Alongside compression metrics, we compute a limited set of descriptive statistics, including normalized diversity measures and repetition-distance indicators, used solely to contextualize structural differences. These quantities are not interpreted as direct estimates of entropy, nor as proxies for semantic quality.

All analyses are applied identically across human-written and machine-generated texts, without relying on model internals, likelihood scores, or semantic representations.

\subsection{Datasets}

We analyze three datasets comparing human-written and machine-generated language across progressively more realistic settings:

\begin{itemize}

\item \textbf{Controlled Human--LLM Corpus.}  
We use the publicly available Human-AI Parallel English Corpus \cite{Reinhart2025Do}. 
For a set of human-written prompts (``Chunk 1''), the dataset provides both human continuations (``Chunk 2'') and model-generated completions produced by six systems: GPT-4o, GPT-4o-mini, Llama 3.1 8B, Llama 3.1 70B, and their Instruct variants. 
This setting enables direct comparison of human and LLM outputs under matched semantic and stylistic constraints.

\item \textbf{Generative Mediation of Knowledge: Wikipedia vs.\ Grokipedia.}  
To study generative mediation within a knowledge infrastructure, we compare 9,279 Wikipedia pages with corresponding Grokipedia entries, where content is rewritten and expanded by a generative system \cite{hadad2026wikipedia}. 
Pages are selected based on high reference counts and revision activity to focus on entries with substantial rewriting. 
Analyses are conducted at the page level, with prefix-based measurements computed at the sentence level.

\item \textbf{Fully Synthetic Social Interaction: Moltbook vs.\ Reddit.}  
We compare Reddit discussions with Moltbook, a platform populated by autonomous LLM agents generating posts and threaded comments \cite{moltbook2026}. 
Reddit data are restricted to pre-2018 posts to minimize the presence of AI-generated content. 
Moltbook data were collected between January 31 and February 10, 2026\footnote{https://moltbook-observatory.sushant.info.np/export}, and Reddit data were obtained via the Pushshift API \cite{baumgartner2020pushshift}.
For each platform, we sample 10,000 posts stratified by length into four categories: \textit{low}, \textit{mid}, \textit{high}, and \textit{very high}, following \cite{DiMarco2024}. This stratified sampling preserves variability in textual length despite the highly skewed distribution of social media content toward short posts. The analyses reported here focus on the central \textit{mid} and \textit{high} length categories, resulting in a final dataset of 20,000 posts for each platform. To ensure comparability across platforms, identical preprocessing steps are applied, including removal of non-textual elements (e.g., URLs, emojis, markup) and normalization of whitespace and encoding. 

\end{itemize}

\section{Results}

\subsection{Controlled Human vs LLM Corpus}
\label{subsec:pnass}

\begin{figure}[!ht]
    \centering
    \includegraphics[width=0.95\linewidth]{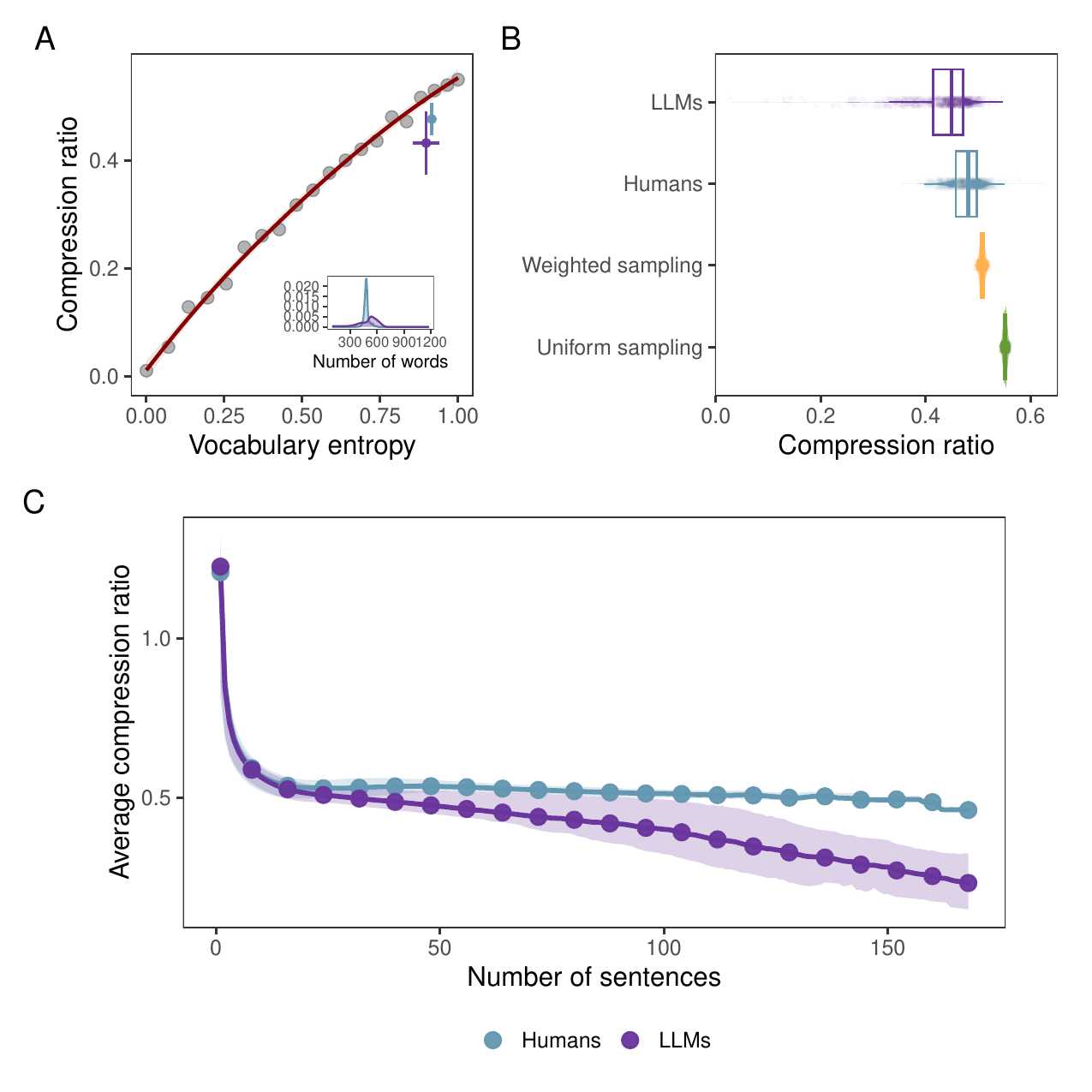}
    \caption{\textbf{(A)} Relationship between vocabulary entropy and compression ratio for texts generated from word distributions with fixed entropy. The colored points show the average values for Humans and LLMs. The bars indicate one standard deviation from the mean. The inset displays the density distribution of document length (number of words) for human-written and LLM-generated texts. \textbf{(B)} Compression ratios distribution for human-written texts, LLM-generated texts, and randomly generated texts. Higher compression ratios correspond to lower compressibility. Each group contains $1000$ documents. \textbf{(C)} Average compression ratio of human-written and LLM-generated texts according to the number of sentences in the text. The shaded area shows the interquartile range of the distribution of compressions. For LLM-generated text, as the length increases, compressibility also increases, unlike human text.}
    \label{fig:entropy_vs_compression}
\end{figure}

In this first section, we introduce the intuition behind using compression as a tool to detect LLM-generated texts.
Our primary goal is to establish an interpretable baseline linking token distributional concentration to compression behavior. To empirically investigate this connection, we rely on the methodology described in Section \ref{subsec:token_distr}. In brief, we consider a parametric family of probability distributions (Eq. \eqref{eq:controlled_entropy_distribution}) whose entropy is controlled by a parameter $h$. We then generate $20$ equally spaced values of $h$ in the interval $\left[\frac{1}{n}, 1\right]$ and generate synthetic texts from the corresponding distributions. For each generated text, we compute the compression ratio (Eq. \eqref{eq:comp_ratio}) at different entropy levels. Additional implementation details are provided in the Methods section.
The results of this analysis are shown in Figure~\ref{fig:entropy_vs_compression}(A), where we observe a positive relationship between these two quantities: higher vocabulary entropy is associated with higher compression ratios, that is, lower compressibility.

Given this result, it is of interest to locate the entropy and compression values obtained by human-written and LLM-generated texts. We do that by considering the Human-AI Parallel English Corpus dataset. Because the compressibility of text is dependent on its length, we first ensure that only human and LLM documents of comparable length are considered. Indeed, as presented in the inset of Figure \ref{fig:entropy_vs_compression}(A), the distributions of words for human and LLM documents in the dataset are considerably different. The former peaks at 479 words, while the latter shows much higher heterogeneity. For this reason, we restrict our analysis to documents between 466 and 489 words, which correspond, respectively, to the first and third quartiles of the distribution for human texts. However, all results are consistent when considering the full dataset.

The colored points and bars in Figure \ref{fig:entropy_vs_compression}(A) locate human and LLMs in terms of average entropy and compression.
While entropy values are broadly comparable across the two groups, more substantial differences emerge in their compression ratios: on average, human-written texts achieve higher compression ratios than LLM-generated ones. 

This pattern is further highlighted in panel B, which displays the distribution of compression ratios in a balanced sample including 1000 documents for each group.
The distributions reveal that LLM-generated texts display lower median compression ratios compared to human-written texts. At the same time, they exhibit a higher heterogeneity, particularly in the left tail of the distribution. 

For reference, we also include a benchmark based on randomly generated documents, each constructed by sampling 479 words, i.e., the mode of the human document length distribution, from the full vocabulary of the dataset, using both uniform and weighted strategies, generating $1000$ documents for each. 
In the first setting, words are selected uniformly at random from the vocabulary. In the second, they are sampled according to the empirical word distribution estimated from the joint corpus of human and LLM texts.

As expected, these random sequences exhibit substantially lower compressibility than both human- and LLM-generated texts, with uniform sampling yielding the highest ratio values. Their near-uniform word usage leads to high entropy and minimal exploitable structure, making them inherently less compressible.

Finally, Panel C reports the differences in compression ratios between human- and LLM-generated texts as document length increases. For each document, we compute the compression ratio incrementally: first considering only the first sentence, then the first two sentences, and so on. We then compute the average and IQR of these distributions, grouping by the number of sentences. As shown in the figure, after roughly 20 sentences an increasingly high textual regularity and redundancy are reflected in the compression ratio curve for LLMs. In contrast, for humans, the metric stabilizes, indicating that human-generated documents do not exhibit increasing regularity as the number of sentences grows; instead, compressibility remains constant throughout the entire text.

\subsubsection{Discriminative Strength of Compression Features}
Building on the compression differences observed above, we examine additional measures that quantify the statistical regularity of the sequence.

\begin{itemize}
    \item \textbf{Conditional compression.} This metric approximates the incremental compression cost of the second half of a document given the first, providing a proxy for context-dependent predictability. It is computed by taking the difference $C(x+y) - C(x)$ and normalizing it by the length in bytes of $y$, i.e., dividing by $|y|$.
    \item \textbf{Average and slope of prefix compression curves.} For each document, we measure the compression ratio as more characters are progressively included. Then, we compute the average over the range, and estimate its linear slope, which characterizes the average compressibility of prefixes and the rate at which structural regularity stabilizes as the sequence grows.
    \item \textbf{Word-order contribution metrics.} To assess the specific contribution of word order to compressibility, we generated a shuffled version of each document by randomly permuting words within sentence boundaries. We then compare the original and shuffled texts using two measures: (i) the compression ratio gap, defined as $R(x_{\text{shuffled}}) - R(x)$, and (ii) the Normalized Compression Distance (NCD) \cite{li2004_ncd}.
    \item \textbf{Normalized character and word entropy.} Complementing compression-based measures, we compute normalized entropy at the character and word levels to measure how evenly they are distributed in the text.
    \item \textbf{Type-Token Ratio (TTR).} We compute the type-token ratio, the proportion of unique words in a text, as a standard measure of lexical diversity and vocabulary richness.
    \item \textbf{Repetition distance features.} For each word, we compute the average and standard deviation of the number of words separating consecutive occurrences. This quantifies the frequency and regularity of repetitions, reflecting structural redundancy.
\end{itemize}

\begin{figure}[t]
    \centering
    \includegraphics[width=0.77\linewidth]{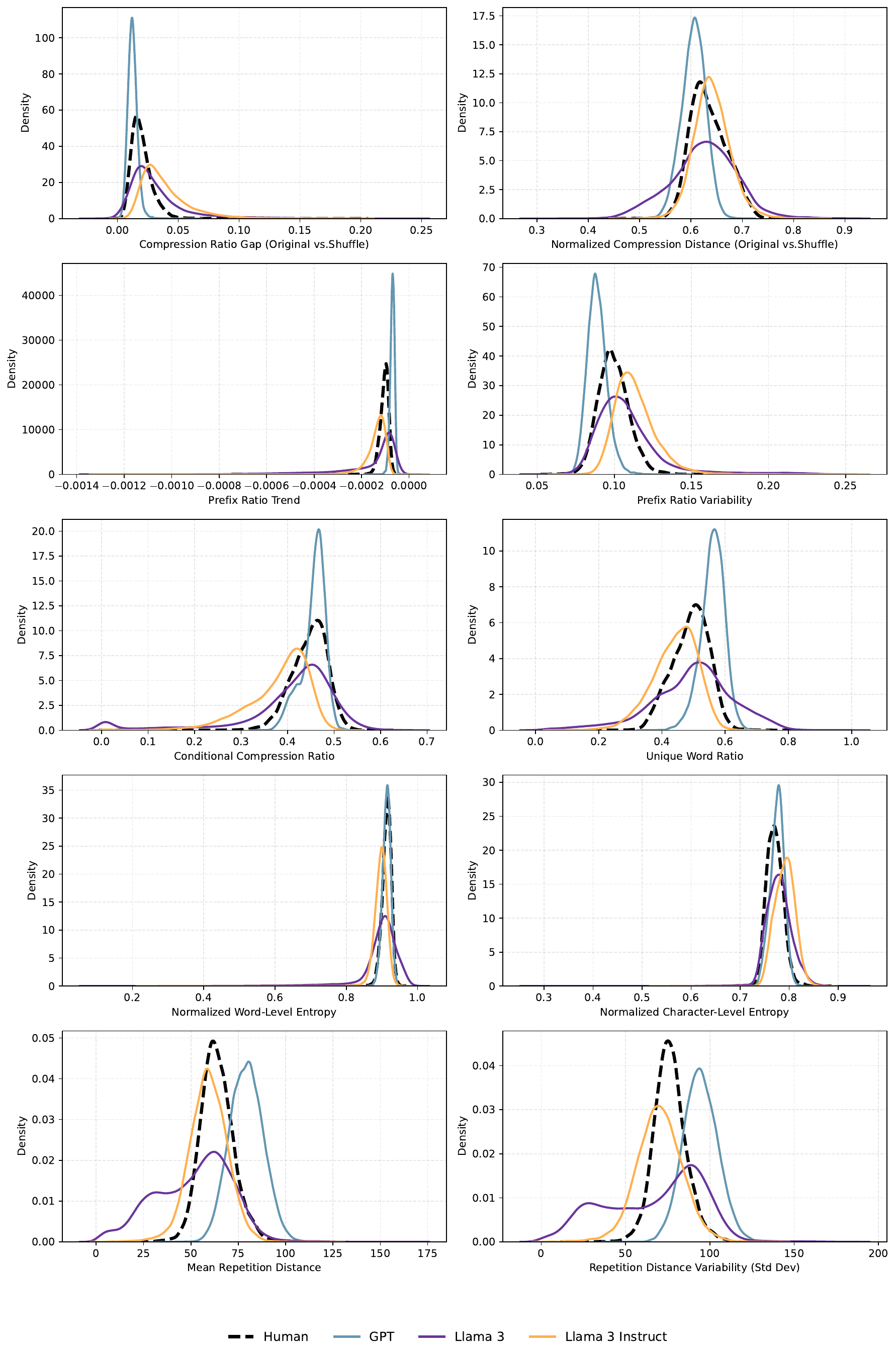}
    \caption{Distribution of structural and compression-based features across human-written and LLM-generated texts in the Human–AI Parallel Corpus.}
    \label{fig:compression_distributions}
\end{figure}

In Figure \ref{fig:compression_distributions}, we show the distribution of these metrics across human and LLM-generated text from the Human-AI Parallel Corpus. This comparison reveals notable differences, even between different model families. GPT models, in particular, display higher conditional compression, lower prefix variability, and sharply peaked distributional profiles, indicating that their sequences stabilize quickly and remain strongly constrained by prior context. Although GPT outputs show comparatively high lexical diversity, they remain more compressible overall, implying that redundancy arises from structural regularity rather than simple word repetition. Llama 3 models occupy an intermediate regime between GPT and human text, while instruction tuning increases variability and shifts distributions closer to the human baseline without eliminating the separation.

To further evaluate the discriminative power of the proposed compression-based features, we trained a Histogram-based Gradient Boosting Classification Tree \cite{scikit_boost} using only the features described above.
We consider three classification settings of increasing granularity. In the most fine-grained setting, we train a seven-class classifier to distinguish between Human, GPT-4o, GPT-4o Mini, Llama 3 70B, Llama 3 70B Instruct, Llama 3 8B, and Llama 3 8B Instruct. The classifier achieves an overall accuracy of $0.65$ (macro F1 = $0.65$). GPT-family models and human text are identified more reliably (F1 scores between $0.75$ and $0.82$), while base Llama models are more frequently confused (F1 scores between $0.46$ and $0.66$). Notably, this performance is comparable to prior detection work \cite{Reinhart2025Do} reporting $66\%$ accuracy using 60 lexical, grammatical, and rhetorical features. This suggests that metrics related to statistical regularity can already capture much of the separable signal.
When collapsing labels into Human, GPT, and Llama, accuracy rises to $0.93$ (macro F1 = $0.90$), indicating that compression features strongly encode family-level generative signatures. Finally, for the binary Human vs.\ LLM task, the classifier achieves $0.93$ accuracy (macro F1 = $0.88$). LLM outputs are detected with very high precision and recall (F1 = $0.96$), while human texts remain the more challenging class (F1 = $0.80$).

\begin{figure}[t]
    \centering
    \includegraphics[width=0.7\linewidth]{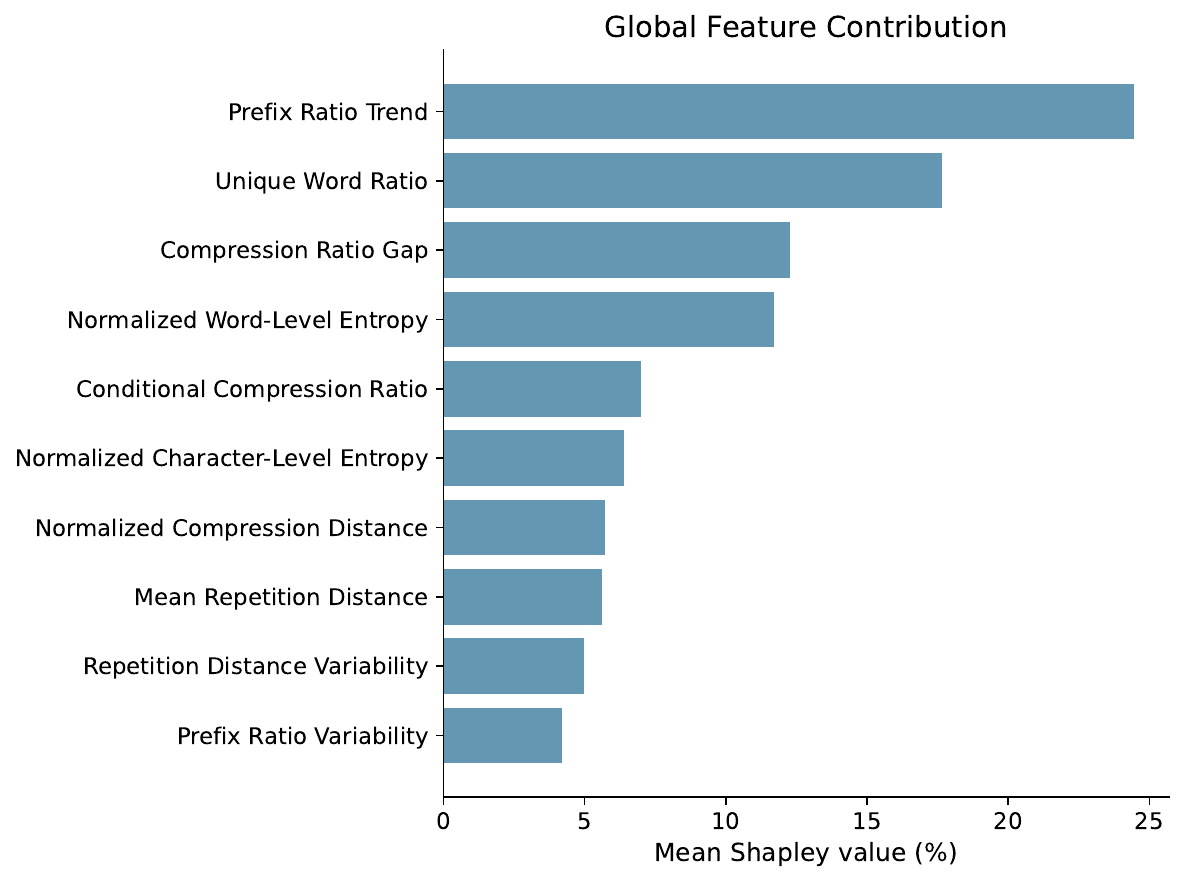}
    \caption{Global feature importance based on mean absolute Shapley values for the Gradient Boosting Classifier. Bars indicate the average magnitude of each feature's contribution to the predicted probability of the Human class across the test set.}
    \label{fig:shapley}
\end{figure}

For model explainability, we perform SHAP analysis on the binary classifier, with the aim of quantifying the impact of each feature on pushing a prediction toward either class. The results, in Fig. \ref{fig:shapley}, reveal that compression-based features and lexical-statistical measures contribute most strongly to the model's decision boundary.

\subsection{Generative Mediation of Knowledge: Wikipedia vs Grokipedia}
\label{subsec:grokipedia}

While the previous section established compression as a proof of concept under controlled generation, here we examine its behavior in a setting where human-authored content is rewritten and expanded by LLMs.

Wikipedia and Grokipedia provide parallel versions of the same encyclopedic content, with Grokipedia consisting of LLM-mediated rewritings of original Wikipedia pages. This setting allows us to examine how compression behaves when human-curated knowledge is partially reshaped through probabilistic generation, and whether such mediation introduces measurable structural differences in the resulting text.

Consistent with the approach used for the Human--AI Parallel Corpus and described in Methods, we start by computing the compression ratio through an incremental prefix analysis, where documents are decomposed into sentences and progressively extended one sentence at a time. Prefix values are aggregated into 20 uniformly spaced bins (minimum 100 observations) and summarized by mean and interquartile range.

\begin{figure}[t]
    \centering
    \includegraphics[width=\linewidth]{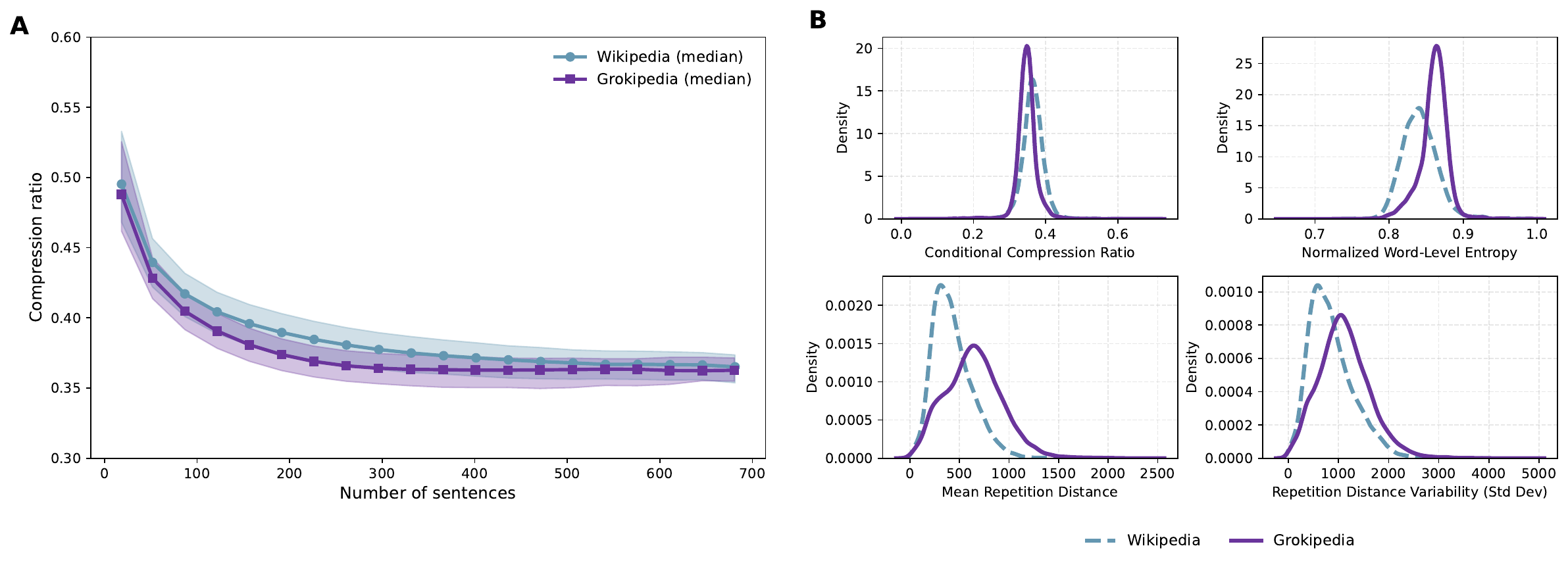}
    \caption{\textbf{(A)} Average compression ratio of Wikipedia and Grokipedia page texts as a function of the number of sentences in the text. The shaded area represents the interquartile range of the compression ratio distribution. \textbf{(B)} Distribution of Conditional Compression Ratio, Normalized Word-Level Entropy, Mean Repetition Distance, and Repetition Distance Variability across texts from Wikipedia and Grokipedia pages.}
    \label{fig:wiki}
\end{figure}

Figure~\ref{fig:wiki} (A) shows that differences in compression ratios begin to emerge after a certain number of sentences.
Around 50 sentences, the interquartile ranges of the two distributions show a diminishing overlap, suggesting that statistical regularities accumulate as text length increases and recurrent structures become more exploitable by compression. As the prefix length approaches 500 sentences, the observed gap progressively narrows. Beyond a certain threshold, this reduction can in part be attributed to the way Grok operates. In particular, Wikipedia pages are predominantly rewritten by Grok in the initial part of the page \cite{triedman2025did}. This portion is also the most frequently accessed by users \cite{Lamprecht2021}, which likely incentivizes more intensive revisions at the beginning of the document. As the prefix extends to include text portions where LLM intervention is comparatively limited, the proportion of human-written text increases and the difference between human and LLM-generated text gradually disappears.
Figure~\ref{fig:wiki} (B) reports the distribution of four relevant metrics. Conditional compression is slightly lower for Grokipedia, indicating reduced incremental compression cost in LLM-mediated text. At the same time, normalized word-level entropy is higher, suggesting broader lexical dispersion. In contrast, Wikipedia texts exhibit more frequent word repetition with lower variance. These patterns differ from those observed in the controlled completion setting, likely reflecting both the longer document structure and different prompting conditions underlying generative mediation. Rewriting human-generated text with LLMs, particularly within an encyclopedic setting, may therefore alter lexical statistics while preserving a lower overall compression rate.

Employing the same model specification and features used for the Human--AI Parallel Corpus, on this dataset, we obtain a binary classification accuracy of $0.85$ (macro F1 = $0.85$). Wikipedia articles are identified with high recall ($0.94$; F1 = $0.86$), while Grokipedia pages reach F1 = $0.84$, with higher precision ($0.93$) but lower recall ($0.76$).

\subsection{Fully Synthetic Social Interaction: Moltbook vs Reddit}
\label{subsec:moltbook}

We extend the analysis to a fully synthetic social setting, comparing human-generated discussions on Reddit with agent-driven interactions on Moltbook, where all content is produced by LLM-based agents. This scenario allows us to examine how compression behaves when conversational dynamics themselves are generated probabilistically rather than emerging from human interaction. Specifically, here we employ the individual post as the unit of analysis. 

\begin{figure}[t]
    \centering
    % Placeholder: Compression ratio distributions (Moltbook vs Reddit)
    \includegraphics[width=\linewidth]{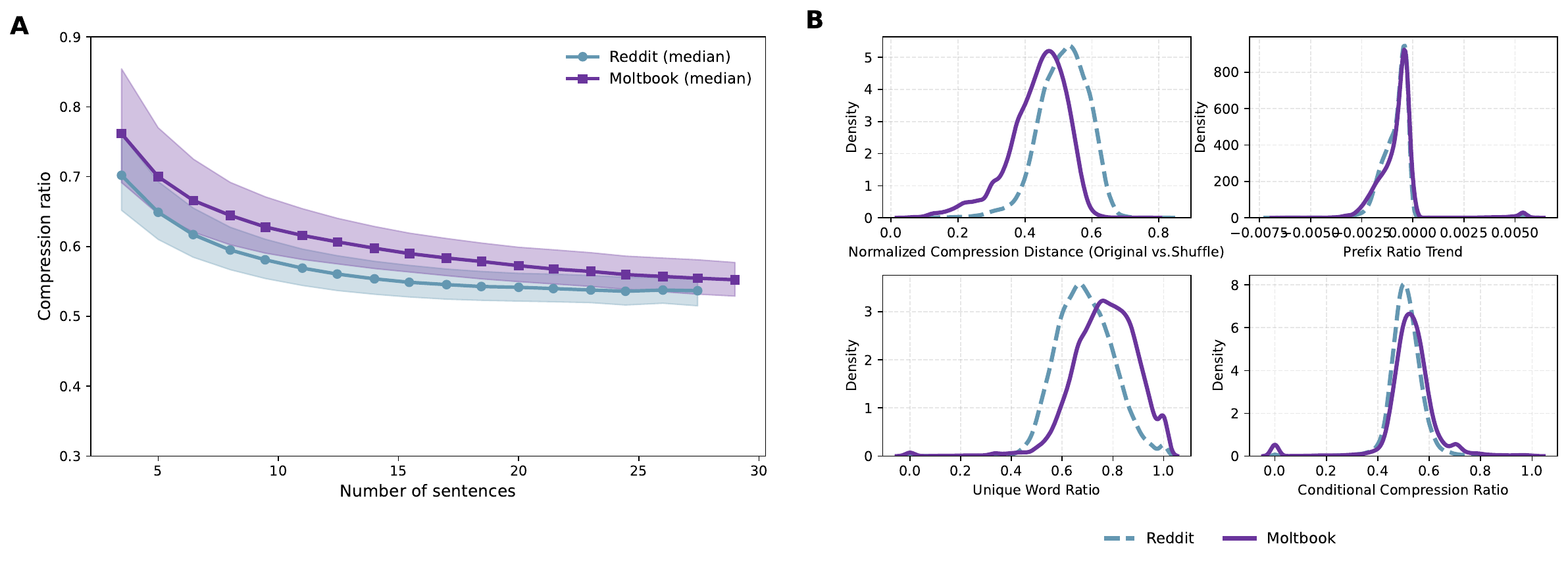}
    \caption{\textbf{(A)} Average compression ratio of Moltbook and Reddit comments as a function of the number of sentences in the text. The shaded area represents the interquartile range of the compression ratio distribution. \textbf{(B)} Distribution of Normalized Compression Distance, Prefix Ratio Trend, Unique Word ratio, and Conditional Compression ratio across Moltbook and Reddit posts.}
    \label{fig:moltbook_reddit_compression}
\end{figure}

Compression ratios are computed following the same procedure described in the Methods and used for the other two datasets. Figure~\ref{fig:moltbook_reddit_compression}(A) reports the average compression ratio as a function of the number of sentences. 

In contrast to the two settings analyzed above, differences between Moltbook and Reddit persist only over a limited range of post lengths, and the gap between the two narrows after approximately 25 sentences. Further, over the shorter range of post lengths considered here, LLM-generated content shows a slightly larger compression ratio, corresponding to lower regularity.

Figure ~\ref{fig:moltbook_reddit_compression}(B) shows that Moltbook interactions exhibit higher lexical diversity, reflected in a rightward shift in unique word ratio, while maintaining comparable overall compressibility. In contrast, Reddit displays higher normalized compression distance and slightly stronger conditional compression, indicating greater sensitivity to sequential ordering and local predictability. The prefix ratio trend remains largely similar across corpora, suggesting comparable incremental redundancy dynamics. 

These results suggest that fully synthetic interaction environments do not simply amplify the structural signatures observed in previous sections. Instead, the emergence of compression-based differences appears to also depend on the scale and structure of the generated text. Since the texts in this case are short and fragmented, we cannot observe the structural regularity that tends to emerge as the length of a text increases.

Furthermore, we should also consider the role of the prompting conditions underlying Moltbook generation. Prompts aimed toward conversational text generation may encourage LLMs to adopt a more casual, context-local, and human-like discourse style that deviates from their baseline generation. We hypothesize that such conditioning attenuates the emergence of systematic long-range regularities that become detectable in extended informational texts, and alters stylistic features in such a way that generated messages effectively show lower regularity and thus compressibility.

To complement the analysis presented above, we apply the same model specification and feature set used for the previous sections. On this task, we obtain a binary classification accuracy of $0.88$ (macro F1 = $0.88$). Performance is highly balanced across classes: Reddit posts are identified with precision and recall $0.88$ (F1 = $0.88$), while Moltbook posts achieve precision $0.88$ and recall $0.87$ (F1 = $0.88$).

\section{Conclusions}

In this work, we examine whether lossless compression can serve as an operational proxy for identifying structural statistical differences between human language and text produced or mediated by large language models. We show that compression-based analysis offers a simple and interpretable way to study how generative systems reshape linguistic structure across tasks and environments.

Across three progressively more complex settings, i.e., controlled continuation tasks, generative mediation of encyclopedic knowledge, and fully synthetic social interaction, we observe that compression consistently captures differences in structural organization between human and LLM-generated language. In controlled continuation tasks, we find that compression reveals a clear separation between human- and LLM-generated text, particularly as a greater number of sentences are included in the analysis. This indicates that probabilistic generation concentrates sequences into more regular and redundant patterns. In generative mediation of encyclopedic content, compression reveals how rewriting reorganizes structure while lexical dispersion increases, highlighting a decoupling between vocabulary diversity and sequential regularity. 
In fully synthetic conversational environments, however, compression-based separation is only observed over a short range of text lengths. Short conversational units and this particular generative regime may limit the accumulation of long-range statistical dependencies.

Our findings indicate that compression captures a structural footprint of plausibility-driven language production. Importantly, this effect should not be interpreted as an evaluation of semantic quality or truthfulness. Compression reflects differences in the generative process rather than differences in meaning. Further, the observed patterns suggest that statistical regularization emerges gradually as generated text becomes longer and more structurally cohesive, rather than appearing uniformly.

As synthetic content becomes increasingly embedded within information ecosystems, understanding these structural shifts is essential for characterizing the dynamics of large-scale language production. Future work may explore how compression-derived signals interact with human post-editing, alternative decoding strategies, and hybrid human–machine production pipelines, as well as their implications for the governance of synthetic information environments.

\bibliography{sn-bibliography}% common bib file

\end{document}